# Real Time Elbow Angle Estimation Using Single RGB Camera

**Muhammad Yahya[1], Jawad Ali Shah[1,*], Arif Warsi[2], Kushsairy Kadir[1], Sheroz Khan[3], M Izani[4]**

1 British Malaysian Institute, Universiti Kuala Lumpur
2 Malaysian Institute of Information Technology, Universiti Kuala Lumpur
3 Dept. of ECE, Faculty of Eng., International Islamic University, Malaysia
4 Visual & Digital Production Department, College of Architecture, Effat University, Jeddah, KSA
* Correspondence: jawad@unikl.edu.my; Tel.: +60-1121215451

**Abstract:** The use of motion capture has increased from last decade in a varied spectrum of applications like film special effects, controlling games and robots, rehabilitation system, animations etc. The current human motion capture techniques use markers, structured environment, and high resolution cameras in a dedicated environment. Because of rapid movement, elbow angle estimation is observed as the most difficult problem in human motion capture system. In this paper, we take elbow angle estimation as our research subject and propose a novel, markerless and cost-effective solution that uses RGB camera for estimating elbow angle in real time using part affinity field. We have recruited five (5) participants of (height, 168 ± 8 cm; mass, 61 ± 17 kg) to perform cup to mouth movement and at the same time measured the angle by both RGB camera and Microsoft Kinect. The experimental results illustrate that markerless and cost-effective RGB camera has a median RMS errors of 3.06° and 0.95° in sagittal and coronal plane respectively as compared to Microsoft Kinect.

---

## 1. Introduction

Human anatomical parts movement are captured and analyzed for various purposes such as controlling games and robots [1], interaction with human machine interface [2], and neurological rehabilitation [3] etc. To track and record the human body parts movement, markers and markerless techniques are used which are then processed using computer. The marker based systems such as Inertial Measurement Units (IMU), Vicon, Qualisys and BTS Smart-D use edges, colors, skin, and wearable gloves etc. as markers while markerless systems are usually based on Ms Kinect [4]. The markers used in most common optical motion capture techniques [5, 6] require controlled environment and placing markers, which are time consuming. These marker based systems have the limitations of mobility, being expensive and difficult to setup [6]. Recently, motion of human body parts have been investigated and analyzed with several different motion capture devices in stroke rehabilitation procedures [7, 8, 11, 12].

There have been numerous attempts to model the human upper limb body segments and measure its orientation with IMU. In [7], researchers have proposed alignment free and self-calibrated method for measuring the elbow angle with initial zero reference pose using inertial sensors. The researchers in [9] proposed wearable sensors that help in assessing the mobility impairment. In [10] the authors have focused on human upper as a link structure with 5 degree of freedom. An unscented Kalman filter has been applied to the data of wearable IMU sensor to estimate the upper arm and forearm movement relationship. However, the movement difference between body segments and the IMU placed on corresponding skin increases significantly with intensive movement. Other works have modeled geometrical constraints in elbow joint and compensate drift by fusing these constraints with particle filter [11]. Inertial sensors have been fused with several other optical motion capture systems like webcam [12] and Ms Kinect [13] and the results are validated with the golden standard system Vicon. In these work the data from optical motion capture system and inertial sensors have been fused to track trajectories and to estimate the range of motion. However, the drifting and placement issues of IMU on human body segments have been reported by many research works [7, 9, 10].

In recent years, researchers have used electromyography to estimate motion of upper extremity. In [14], the elbow joint angle estimation has been used to tele-operate a robot based on Surface Electromyography (sEMG) signals. The estimation of elbow joint motion has been accomplished using auto regressive moving average. The authors in [15] classified the shoulder joint motion into 4 classes by quadratic discriminant analysis of EMG signals. In [16], the authors have estimated 4 DoF across upper limb, shoulder and elbow joints with simultaneous and continuous kinematics. Their results show that independent component analysis with single artificial neural network (ANN) for multiple-joint kinematics estimation is effective and feasible. Another technique for elbow angle estimation has been stated by authors in [17]. In their work, the authors estimated elbow joint with goniometer calibration and trained the developed ANN to measure the elbow joint angle from EMG signals. The EMG signals pattern are classified and recognized according to human motion to estimate elbow joint angle using neural network by many other researchers [18, 19]. The elbow joint angle estimation with sEMG signals depends on the electrodes position as well as the length of muscles (Biceps and triceps) during force [20].

Single RGB camera has also been used to estimate the human motion using markers. In [21], the patients gait motion has been analyzed with RGB camera with bulls-eye marker placed on neck, shoulder, waist, elbow and wrist. The markers are then tracked by the camera while the patient taking a cup of tea from desk to mouth and placing it back on the desk.

Most of the marker based motion estimation systems are time consuming, expensive and require calibrated and well structure environment. In this paper, we present a cost-effective markerless system to estimate elbow joint angle based on bottom-up approach in [22]. The proposed system uses a single RGB camera that requires less computational power without any calibrated environment. We have released the code for estimating the elbow angle in real time for a single person.

## 2. Neural Network Model

In [22], the authors proposed a bottom-up approach to estimate 2D poses of multiple human in a real-time scene. The human body segments are located and associated by means of two branches of same chronological prediction process. We have used this trained model to estimate the elbow angle for a single person using only one RGB camera.

A colour image of size $x \times y$ is given to the system as input, which produces the 2D locations of human body segments. Feed-forward network concurrently predicts a set of confidence maps $\mathbf{P}$ of body segment 2D locations and a set $\mathbf{Q}$ of part affinities field which expresses the amount of association between segments.

The set $\mathbf{P} = (\mathbf{P}_1, \mathbf{P}_2, \ldots, \mathbf{P}_G)$ and the set $\mathbf{Q} = (\mathbf{Q}_1, \mathbf{Q}_2, \ldots, \mathbf{Q}_H)$ have $G$ confidence map and $H$ vector field respectively one per segment where $\mathbf{P}_g \in \mathbb{R}^{x \times y}$ and $\mathbf{Q}_h \in \mathbb{R}^{x \times y \times 2}$. The neural network has been divided into two branches that makes iterative predictions over successive time values of $k \in \{1, 2, \ldots, K\}$ for confidence map $\mathbf{P}^k$ and affinity fields $\mathbf{Q}^k$.

The first ten layers of convolutional neural network in [22], first analyze the image and then produce a set of feature maps $\mathbf{F}$. These features maps are given as input to the first stage of each branch. A set of confidence maps $\mathbf{P}^1 = \alpha^1(\mathbf{F})$ and a set of $\mathbf{Q}^1 = \beta^1(\mathbf{F})$ affinity fields are produced by network at first stage. Where $\alpha^1$ and $\beta^1$ are the CNNs for inference at Stage 1. The previous stage predictions and image features $\mathbf{F}$ from both branches are combined in each subsequent stage and produce refine predictions.

$$\mathbf{P}^k = \alpha^k(\mathbf{F}, \mathbf{P}^{k-1}, \mathbf{Q}^{k-1}), \forall k \geq 2 \qquad (1)$$

$$\mathbf{Q}^k = \beta^k(\mathbf{F}, \mathbf{P}^{k-1}, \mathbf{Q}^{k-1}), \forall k \geq 2 \qquad (2)$$

Where $\alpha^k$ and $\beta^k$ are CNNs for inference at stage $k$.

Two loss function one at each stage, are applied at the end of each stage to iteratively predict confidence maps in first branch and affinity fields of human key points in second branch. Specifically, at both branches the loss function at stage $k$ is

$$l_{\mathbf{P}}^k = \sum_{g=1}^{G} \sum_{\mathbf{p}} \mathbf{D}(\mathbf{p}) . \parallel \mathbf{P}_g^k(\mathbf{p}) - \mathbf{P}_g^*(\mathbf{p}) \parallel_2^2 \qquad (3)$$

$$l_{\mathbf{Q}}^k = \sum_{h=1}^{H} \sum_{\mathbf{p}} \mathbf{D}(\mathbf{p}) . \parallel \mathbf{Q}_h^k(\mathbf{p}) - \mathbf{Q}_h^*(\mathbf{p}) \parallel_2^2 \qquad (4)$$

Where $\mathbf{P}_g^*$ and $\mathbf{Q}_h^*$ are the ground truth confidence maps and part affinity vector fields respectively generated from the annotated 2D key points, $\mathbf{D}$ is a binary mask with $\mathbf{D}(\mathbf{p}) = 0$, when at location $\mathbf{p}$ annotations are missing. The overall objective is

$$l = \sum_{k=1}^{K} (l_{\mathbf{P}}^k + l_{\mathbf{Q}}^k) \qquad (5)$$

The confidence maps for particular body parts are generated as per equation (6). Let $\mathbf{i}_g \in \mathbb{R}^2$ is body segments $g$ ground truth position. At location $\mathbf{p} \in \mathbb{R}^2$ in $\mathbf{P}_g^*$

$$\mathbf{P}_g^*(\mathbf{p}) = \exp\left(-\frac{\parallel \mathbf{p} - \mathbf{i}_g \parallel_2^2}{\sigma^2}\right) \qquad (6)$$

Let $\mathbf{i}_{g_1}$ and $\mathbf{i}_{g_2}$ are the body segments $g_1$ and $g_2$ groundtruth positions from the limb $h$. At image point $\mathbf{p}$, the vector field $\mathbf{Q}_h^*$ are define as

$$\mathbf{Q}_h^*(\mathbf{p}) = \begin{cases} \mathbf{v} & \text{if } \mathbf{p} \text{ on limb } h \\ \mathbf{0} & \text{otherwise} \end{cases} \qquad (7)$$

Where $\mathbf{v} = (\mathbf{i}_{g_2} - \mathbf{i}_{g_1})/\parallel \mathbf{i}_{g_2} - \mathbf{i}_{g_1} \parallel_2$, is a unit vector in the limb direction. Those points which are within a distance defined by the following line segment are said to be on the limb.

$$0 \leq (\mathbf{p} - \mathbf{i}_{g_1}) . \mathbf{v} \leq r_g \text{ And } \left| (\mathbf{p} - \mathbf{i}_{g_1}) . \mathbf{v}_\perp \right| \leq \sigma_r$$

Here $\sigma_r$ (limb width) is the distance in pixels $r_g = \parallel \mathbf{i}_{g_2} - \mathbf{i}_{g_1} \parallel_2$ and $\mathbf{v}_\perp$ is a vector perpendicular to $\mathbf{v}$.

The association between the body segments is computed by the line integral over the matching part affinity field, along the line segment connecting body part locations. Specifically, for two body segments locations $\mathbf{s}_{g_1}$ and $\mathbf{s}_{g_2}$, the predicted vector fields $\mathbf{Q}_h$ are sampled along the line segment to estimate the confidence in their association.

$$Z = \int_{a=0}^{a=1} \mathbf{Q}_h(\mathbf{p}(a)) . \frac{\mathbf{s}_{g_2} - \mathbf{s}_{g_1}}{\parallel \mathbf{s}_{g_2} - \mathbf{s}_{g_1} \parallel_2} da \qquad (8)$$

Where $\mathbf{p}(a)$ interpolates the position of the two body segments $\mathbf{s}_{g_1}$ and $\mathbf{s}_{g_2}$.

Equation 8 defines the score of each body limb using the line integral calculation. Thus, we obtain a set of body part detection for a single person.

## 3. Elbow Angle Calculation

The image captured through RGB camera is passed through trained model, the confidence map for upper limb joints locations are estimated and are associated through part affinity filed. After confidence map estimation and vector field association, the three joints coordinates are extracted for angle estimation as shown in Figure-1 (c).

The Elbow angle is estimated with RGB camera and its results are then compared with Microsoft Kinect in Results section.

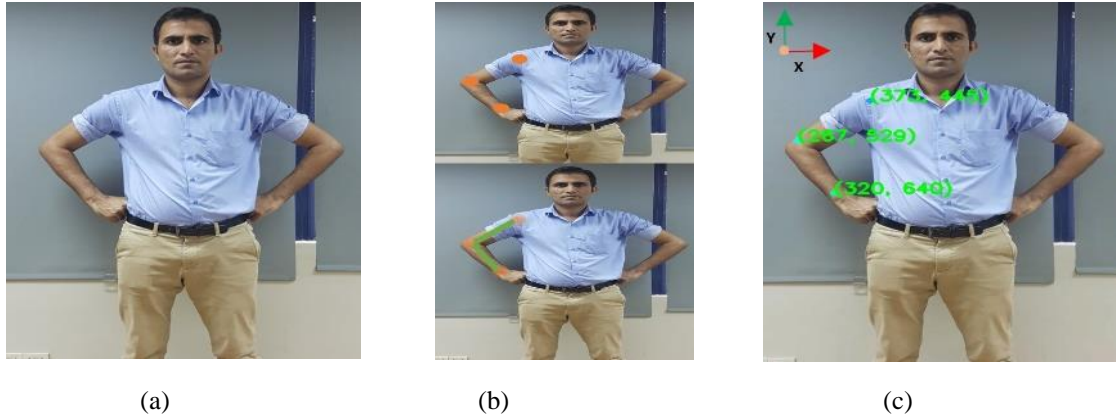

(a)                    (b)                    (c)

**Figure-1** Shows overall process. (a) Input image, (b) Two-branch CNN to predict confidence maps for upper limb segments and vector fields for segments association. (c) Joints location coordinates are extracted and are used to estimate elbow angle

## 4. Methodology

Figure-2 shows the overall process from acquisition to angle estimation for both motion capture system. The camera sends thirty frames per second to the trained neural network model. Two branches of neural networks predict the confidence map and vector field simultaneously in a single layer. Once the body parts 2D position and orientation are localized, then the coordinates of the human anatomical parts are extracted from the frames using body's center of mass as region. The coordinates of shoulder, elbow and wrist of right limb respectively are processed for angle calculation. The elbow angle is estimated from the vector transformation of these coordinates using the following formula.

$$\boldsymbol{u} = \boldsymbol{x_s} - \boldsymbol{x_e}$$

$$\boldsymbol{v} = \boldsymbol{x_w} - \boldsymbol{x_e}$$

$$\theta = \cos^{-1}\left(\frac{<\boldsymbol{u}.\boldsymbol{v}>}{\parallel \boldsymbol{u} \parallel \parallel \boldsymbol{v} \parallel}\right)$$

Where $\boldsymbol{x_s}$, $\boldsymbol{x_e}$ and $\boldsymbol{x_w}$ are the coordinate vectors for shoulder, elbow and wrist joints respectively. While $<.,.>$ denotes the inner product and $\parallel . \parallel$ is the $l_2$ norm of the vector.

Microsoft Kinect acquires 3D depth images by using built-in infrared projector and complementary metal oxide semiconductor sensor to track human body-joint motion in real time [23]. We use the same mathematical formula for estimating elbow joint angle with Kinect as for RGB camera.

Microsoft Kinect tracks the human skeleton with a 30 frames per second while the RGB camera is using 15 frames per second. We synchronized the frames from both system with respect to time in milliseconds. So it synchronizes all those frames of RGB camera which are having matching time with that of Microsoft Kinect frames. The pseudo code for synchronizing RGB camera with Microsoft Kinect based on time matching frames is as follow,

1.  **START**
2.  **INPUT:**    KDataCollection, Kinect angle data w.r.t time
3.                     RGBDataCllection, RGB camera angle data w.r.t time
4.  **OUTPUT:**    synDataCollection, Merged Data for Kinect and RGB angles w.r.t to time
5.  **INITIALIZE:** KinectTime, store current Kinect time
6.                     RGBTime, store current RGBCamera time
7.  **FOREACH** Kdata in KDataCollection
8.    **SET** KinectTime = Kdata [Time]
9.      **FOREACH** RGBData in RGBDataCllection
10.      **SET** RGBTime = RGBData [Time]
11.        **IF** KinectTime = RGBTime **THEN**
12.          synDataCollection.Add (KinectTime, Kdata[Angle],RGBData [Angle])
13.        **END IF**
14.      **END FOREACH**
15.  **END FOREACH**
16.  **RETURN** synDataCollection
17.  **STOP**

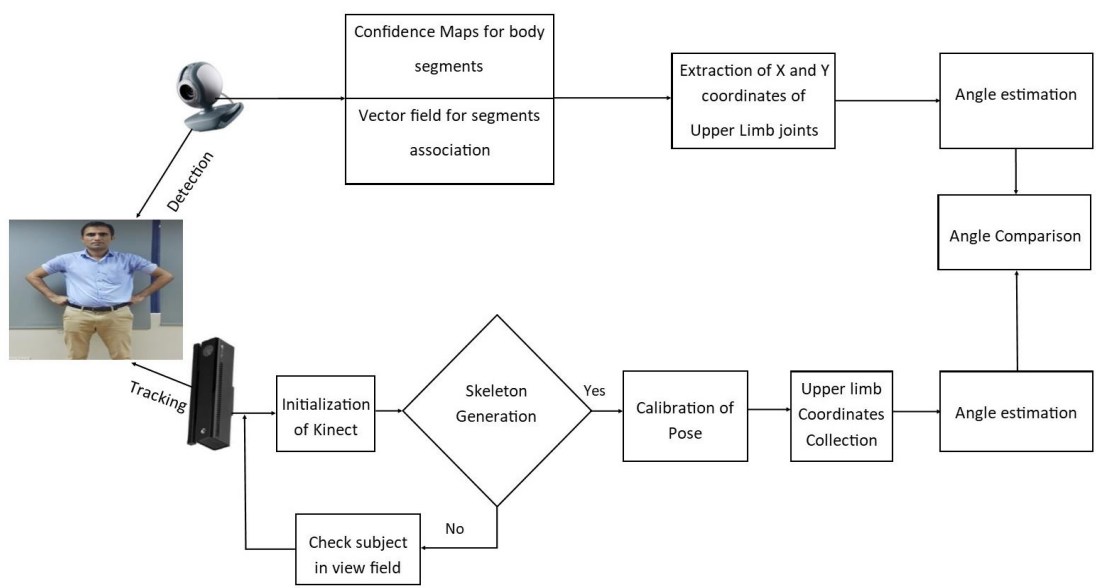

**Figure-2** overall process for Estimation of Elbow angle

### 4.1 participants

For angle estimation, we performed cup to mouth action on five participants (height, 168 ± 8 cm; mass, 61±17 kg). The participants take part voluntarily after taking ethical approval from Unikl Research Ethics Committee. Written consent was obtained before data acquisition and all participants were having no injury.

### 4.2. Experimental setup

In our experiment, we engaged five healthy subject and didn't use any markers for tracking the movement of subject's body. A single experiment consists of three trails for each subject. During each trail, the subject was seated on a chair and asked to perform the experiment by lifting a cup from table and moving it near to the mouth. The cup was then brought back it to its initial position. During the entire trial, both RGB camera as well as Microsoft Kinect track the joints coordinates in real time to compute the estimated angles independently in Sagittal and coronal planes.

The experimental design reflects the movement related to the daily activities. In rehabilitation systems, this task plays an important role to track the motion in degrees where therapist can gauge patient performance based on the data acquired.

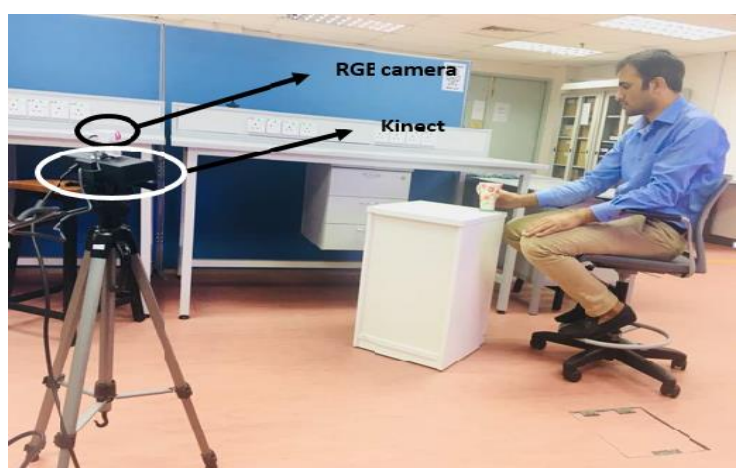

**Figure-3.** Experimental setup

## 5. Software setup

Microsoft Kinect was programmed using C# with Coding4fun toolkit [24] to estimate the elbow joint angle. The angle estimation for the frames captured through RGB camera was done using python. Both of these motion capture systems were used to track human body simultaneously in real time. However, Microsoft Kinect acquires three coordinates while RGB camera acquires two coordinates for shoulder, elbow and wrist joint. Similarly, the Microsoft Kinect has a frame rate of 30 frame per second whereas RGB camera has a frame rate of 15 frame per second. To adjust the frame rates, the frames of RGB camera are synchronized with Microsoft Kinect with respect to time (within millisecond accuracy) by down sampling Microsoft Kinect frame rate. To get an accurate comparison of the two motion capture systems, the analysis has been done using the results after synchronization.

## 5. Result and Discussion

Initially the subject was asked to make a static pose in two different planes i.e. sagittal and coronal to compare the accuracy of elbow joint angle measured through RGB camera and MS Kinect. It can be seen from figure-4 that the angles estimated through both systems are nearly equal. The results depicted in fig-4 (a) shows the accuracy of the angles computed through RGB camera and Ms Kinect in the sagittal plane which are 152.27° and 150° respectively. In a separate experiment, carried out in the coronal plane, the angle recorded by RGB camera is 38.55° which is again approximately equal to that measured by MS Kinect i.e.  39°.

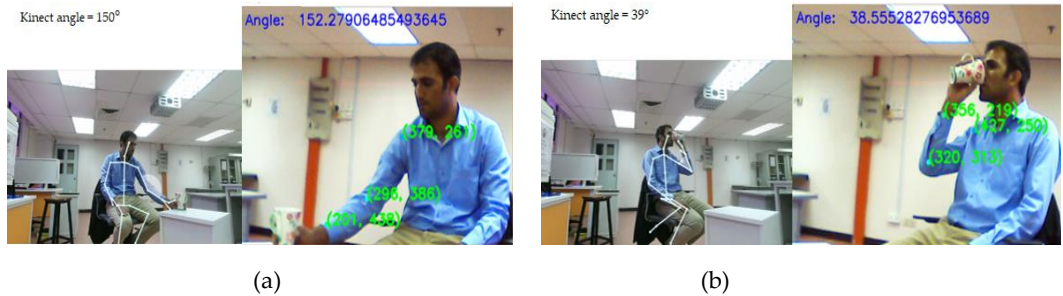

(a)                                              (b)

**Figure-4.** Shows angle estimation in two different static poses by RGB camera and Ms Kinect. The angle in black color on left side of both (a) and (b) is of Ms Kinect and angle in blue color is of RGB camera.

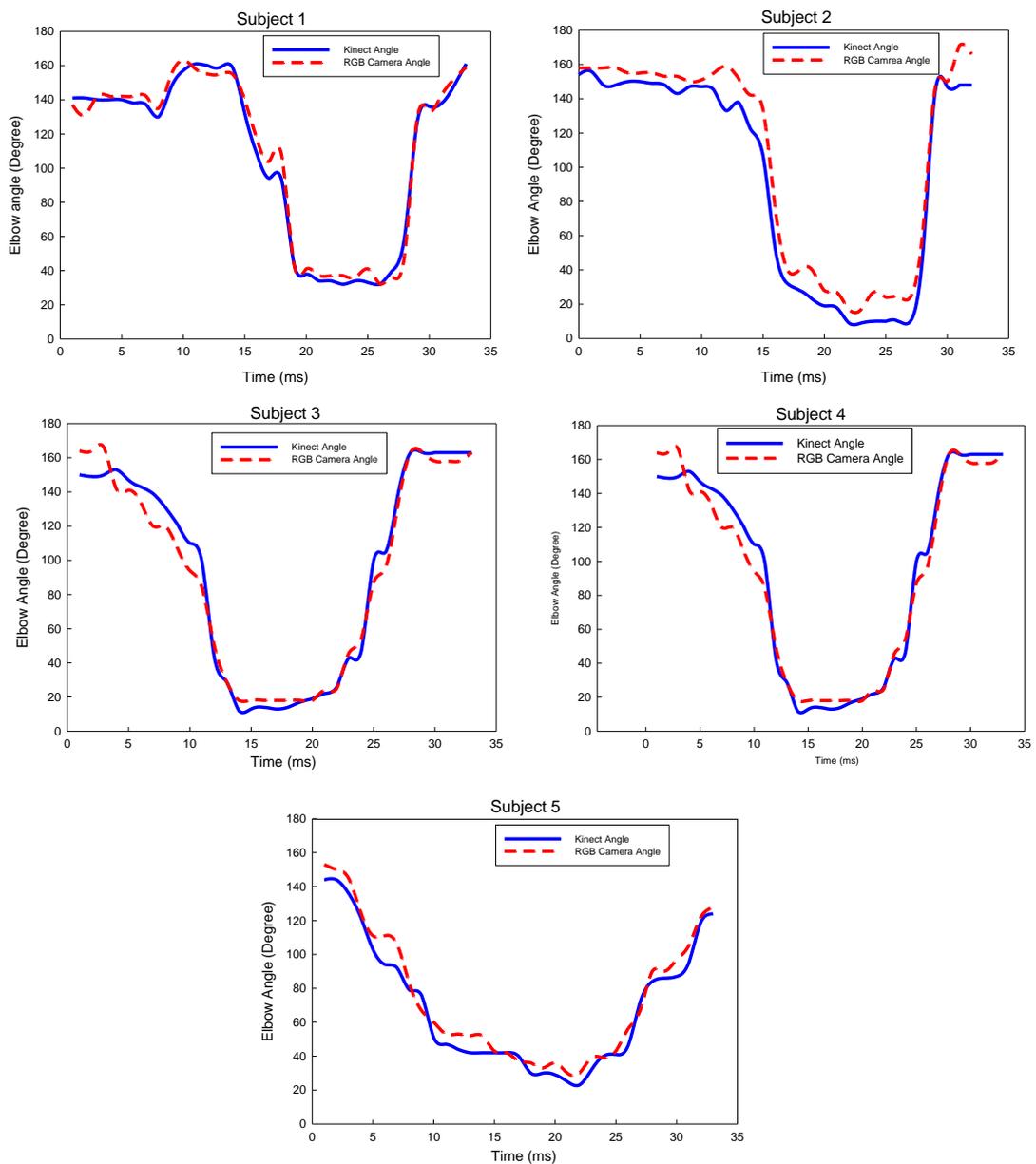

**Figure-5**. Elbow angle measured in degree by RGB camera and Microsoft Kinect in sagittal plane.

After the static pose trail, the subject was asked to perform cup to mouth experiment in both Sagittal and coronal planes to record the motion trajectory. Figure-5 shows the results of elbow joint angle estimated for each of the subject in the Sagittal plane. The motion trajectory shown is based on a single trial for the timely synchronized frames of both motion capturing systems. It can be seen from the results that the elbow joint angles calculated during motion with RGB camera nearly follow the results of MS Kinect for all the subjects.

The motion trajectories of elbow joint angle performing the same task in coronal plane are shown in Figure-6. It is evident that the results in coronal plane are quite promising as compared to those in sagittal plane (Figure-5) due to precise position detection of the wrist joints coordinates.

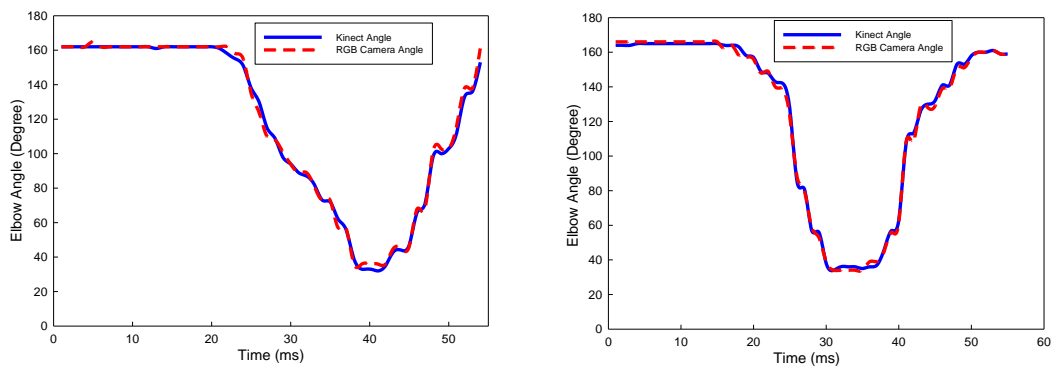

**Figure-6.** Estimates elbow angle in degrees with RGB camera and Microsoft Kinect in coronal plane for two trails.

Figure-7 illustrates error histograms from the angle estimated by RGB camera and Microsoft Kinect in sagittal and coronal plane. The errors are computed after synchronizing the frames of RGB camera and Microsoft Kinect. The error range in sagittal plane is between $\pm10°$ due to imprecise detection of the wrist joint. While, in coronal plane wrist joint is precisely detected and showing the lesser range of error. The median of Root Mean Square Error (RMSE) for sagittal and coronal planes is given in table-1 for all the subjects.

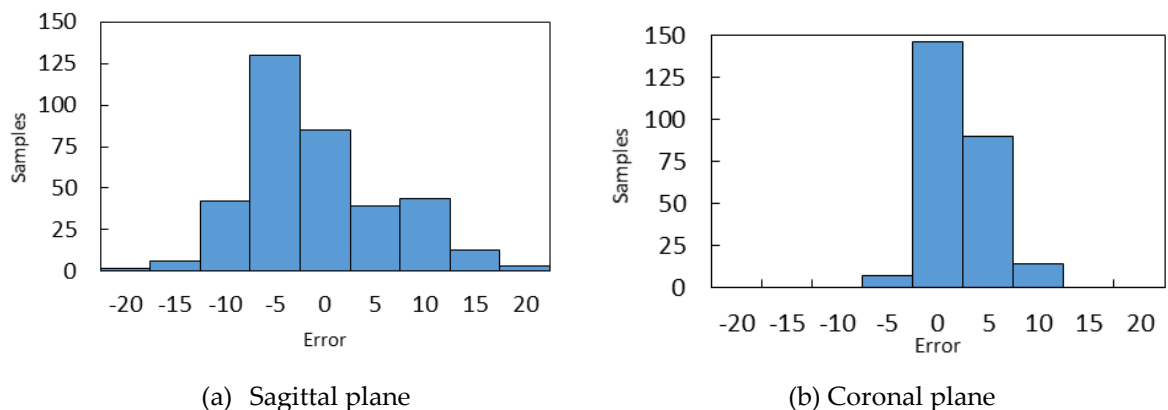

(a) Sagittal plane           (b) Coronal plane

**Figure-7.** Error histogram of RGB camera and Microsoft Kinect

Experimental results show that the use of a single RGB camera can be a cost-effective alternative solution to Microsoft Kinect for estimating joint elbow angles in applications like rehabilitation etc.

**Table-1.** RMSE in frames of RGB camera w.r.t Microsoft Kinect in sagittal and coronal plane

| Subject | Sagittal plane(RMSE) | Coronal Plane(RMSE) |
|---|---|---|
| 1 | 1.5° | 1° |
| 2 | 4° | 2° |
| 3 | 2° | 0.5° |
| 4 | 2.5° | 0.5° |
| 5 | 3.5° | 1° |

## 6. Conclusions

The comparison of RGB camera with Microsoft Kinect is a key aspect for the development of low cost markerless motion capture system. This paper illustrates the exploration and evaluation of tracking motion accuracy amongst low cost RGB camera and Microsoft Kinect. Markerless RGB camera based motion capture system has shown reasonably accurate results in coronal plane and can be used to estimate join elbow angle in rehabilitation.

Future studies should address the wrist joint position detection issues in sagittal plane. The proposed work can be used in other areas for movement-based experiments such as motion analysis of athletes, posture estimation, security camera and many more.

**Acknowledgments:** This research has been generously supported by a research grant by the Ministry of Science, Technology and Innovation (MOSTI) under the Program Flagship DSTIN for the development of a new technology identified as "Building Our Robotic Competitiveness in Medical Healthcare: Development of Robots for Assisted Recovery and Rehabilitation".

**Author Contributions:** Muhammad Yahya raised up the idea and wrote the paper. Jawad Ali Shah supervised the whole process. Arif Warsi helped to develop the code. The results were discussed, validated and modified by Kushsairy Kadir. Sheroz khan and Izani review the manuscript and modified the English.

**Conflicts of Interest:** I declare that there is no conflict with other research work.